\documentclass[letterpaper, 10 pt, conference]{ieeeconf}
\usepackage{url}
\usepackage{graphicx}
\usepackage{amsfonts}
\usepackage{fancyhdr}
\usepackage{comment}
\usepackage{times}
\usepackage{amsmath}
\usepackage{changepage}
\usepackage{cite}
\usepackage{amsmath,amssymb,amsfonts}
\usepackage{algorithmic}
\usepackage{graphicx}
\usepackage{textcomp}
\usepackage{xcolor}
\usepackage{mathrsfs}
\usepackage{balance}

\usepackage{graphicx, subfig} 
\usepackage{algorithm}
\IEEEoverridecommandlockouts                       
\def\BibTeX{{\rm B\kern-.05em{\sc i\kern-.025em b}\kern-.08em
    T\kern-.1667em\lower.7ex\hbox{E}\kern-.125emX}}
\begin{document}

\title{\LARGE \bf Robust Lane Marking Detection Algorithm Using Drivable Area Segmentation and Extended SLT}

\author{
Umar Ozgunalp, Rui Fan, Shanshan Cheng, Yuxiang Sun, Weixun Zuo,\\\ \ \ \ \  Yilong Zhu, Bohuan Xue, Linwei Zheng, Qing Liang, Ming Liu
\thanks{U. Ozgunalp is with the Electrical and Electronics Engineering, Cyprus International University, North Cyprus, Turkey. Email: umarozgunalp@gmail.com}
\thanks{R. Fan is with the Department of Electronic and Computer Engineering, the Hong Kong University of Science and Technology, Hong Kong SAR, China, as well as Hangzhou ATG Intelligent Technology Co. Ltd., Hangzhou, China. Email: eeruifan@ust.hk}
\thanks{S. Cheng is with Guolu Gaoke Engineering Technology Institute Co. Ltd., Beijing, China. Email: cpms2001@163.com}
\thanks{Y. Sun, W. Zuo, Y. Zhu, B. Xue, L. Zheng, Q. Liang and M. Liu are with the Robotics and Multiperception Laboratory, Robotics Institute, the Hong Kong University of Science and Technology, Hong Kong SAR, China. Emails:  \{eeyxsun, eezuo, yzhubr, bxueaa, lzhengad, qliangah, eelium\}@ust.hk}
}


\maketitle
\begin{abstract}
In this paper, a robust lane detection algorithm is proposed, where the vertical road profile of the road is estimated using dynamic programming from the v-disparity map and, based on the estimated profile, the road area is segmented. Since the lane markings are on the road area and any feature point above the ground will be a noise source for the lane detection, a mask is created for the road area to remove some of the noise for lane detection. The estimated mask is multiplied by the lane feature map in a bird's eye view (BEV). The lane feature points are extracted by using an extended version of symmetrical local threshold (SLT), which not only considers  dark light dark transition (DLD) of the lane markings, like (SLT), but also considers parallelism on the lane marking borders. The segmentation then uses only the feature points that are on the road area. A maximum of two linear lane markings are detected using an efficient 1D Hough transform. Then, the detected linear lane markings are used to create a region of interest (ROI) for parabolic lane detection. Finally, based on the estimated region of interest, parabolic lane models are fitted using robust fitting. Due to the robust lane feature extraction and road area segmentation, the proposed algorithm robustly detects lane markings and achieves lane marking detection with an accuracy of $\boldsymbol{91\%}$ when tested on a sequence from the KITTI dataset.

\end{abstract}

\section{Introduction}
\label{sec.introduction}
Advanced driving assistance systems (ADAS) are becoming an essential component for intelligent vehicles, with many commercial car manufacturers already including such systems on their recent models \cite{Fan2019}. According to Euro NCAP, which supplies vehicle safety rating system, a vehicle model needs to be equipped with a robust crash avoidance technology to be able to get a ``5-star safety" rating. An ADAS includes many functionalities, such as autonomous emergency breaking (AEB) for pedestrians, vehicle state estimation, lane departure warning system \cite{Bertozzi1998}, etc.

Lane detection is vital and essential for the lane departure warning system or cruise control \cite{fan2018real}.
A lane detection system usually consists of two components: feature extraction and lane modeling \cite{Bertozzi1998, Fan2018thesis}, where feature extraction algorithms segment the road area and extract the lane features \cite{neto2013real}, while the lane modeling algorithms 
fit lanes to a model (mathematical equation), based on the assumptions, such as constant road width \cite{Bertozzi1998}, parallel lane markings \cite{schreiber2005single}, flat road \cite{ozgunalp2015lane}, and constant lane marking width \cite{Bertozzi1998}. The noise sources for a lane detection algorithm are either located above the road area, caused by cars, trees, road signs, etc., or located on the road, e.g., shadows, road markings (e.g., arrows), and cracks. Due to the above-mentioned noise sources, and saturation on the image, detecting lanes is still a challenging task. Thus, a robust feature extraction algorithm is necessary.
\subsection{Related Work}
Lane detection algorithms are generally applied in two domains: image domain and bird's eye view (BEV) domain. In the image domain, one important road image property is vanishing point (VP). Due to the perspective mapping, when the lanes are parallel to each other and the curvature of the road is limited, they converge and intersect at a VP \cite{ma2018multiple}. The VP was used by many algorithms \cite{ozgunalp2016multiple} to improve the robustness of the applied algorithm. In \cite{schreiber2005single}, the VP was first estimated and then, considering all the lanes need to intersect at the VP (one parameter is already known), the lanes are detected efficiently and robustly using the 1D Hough transform. On the other hand, an input image is generally transformed into a BEV using inverse perspective mapping (IPM). IPM assumes that the road is flat and the intrinsic and  extrinsic parameters are known. Based on these assumptions, a BEV image can be created. In the BEV domain, lanes are generally parallel to each other (except converging and diverging lanes). In our previous paper \cite{ozgunalp2015lane}, a histogram of the orientation for the lane feature point was estimated under the parallel lane assumption. Thus, a global orientation for the lanes was calculated. This approach resembles detecting the VP first, and then, detecting lanes based on the detected VP.

In the literature, many feature extractors have been proposed for lane detection, including edge detectors \cite{kluge1995deformable}, the local threshold \cite{broggi2010terramax}, the symmetrical local threshold (SLT) \cite{veit2008evaluation}, ridege detector \cite{lopez2010robust}, Otsu algorithm \cite{otsu1979threshold}, Gabor filters \cite{mccall2006video} and the top hat filter \cite{aubert1991autonomous}. Although there is no benchmark for lane feature extraction, the authors of \cite{veit2008evaluation} tested the most common lane feature extractors using the ROMA dataset. In spite of  its computational efficiency, SLT achieved best accuracy among those tested. The SLT utilizes the fact that the lane markings have a dark-light-dark transition property. However, in \cite{ozgunalp2014robust}, we proposed an extension to the SLT, in which the proposed feature extractor uses only the DLD property of the lane markings but also the fact that the lane borders must be parallel to each other, thus, achieving better accuracy when tested on the ROMA dataset. Furthermore, an orientation for each feature point was calculated, which is not possible with the conventional SLT.
The estimated feature map was then filtered using the segmented road area estimated by applying dynamic programming on the v-disparity map. Thus, noise sources above the road were removed and noise due to cars, trees, buildings and sky were eliminated. 

The second step of a lane detection algorithm is lane modeling. Many lane models have been introduced. The linear lane model \cite{schreiber2005single}, parabolic lane model, linear-parabolic lane model \cite{lim2009lane}, clothoid lane model \cite{loose2009kalman}, and splines \cite{wang2004lane} are some of the commonly used models. The linear lane model, defines lanes with as linear line and is suitable for high-speed roads where the curvature of the road is limited. Furthermore, due to the limited number of parameters, and depending on the optimization algorithms adopted, using the linear lane model can be more computationally efficient, and it is hence, preferred for systems with limited computational power. On the other hand, the parabolic lane model can define lanes with a constant curvature. In \cite{lim2009lane}, a linear-parabolic lane model was introduced, where the near field was defined using a linear model and the far field was defined using a parabola. In \cite{wang2004lane}, splines are introduced. Depending on the number of control points used, splines can define any arbitrary shape. 
However, more flexible models can also be more sensitive to noise. For instance, parabolic lane model restricts the lane shape to a constant curvature, which is generally true. 
Furthermore, estimating more parameters generally need more computation. The parameters of the lane model needs to be estimated using an optimization algorithm. 
For lane detection, RANSAC \cite{borkar2009robust}, the Hough transform \cite{son2015real}, and dynamic programming \cite{ozgunalp2016multiple} are some of the optimization algorithms used.

\subsection{Contributions}
In this paper, a hybrid approach is proposed. 
In the first step, the vertical road profile is estimated using dynamic programming on a v-disparity map and the road area is segmented.  Thus, a substantial amount of noise is eliminated by segmenting the rest of the feature points. In the second step, the input image is converted to the BEV image using IPM and the lanes are detected using only the feature points that appear on the road. 
In this paper, IPM is applied to both input image and the segmented road area to estimate BEV, and lanes are detected in this domain. 
\subsection{Paper Structure}

The remainder of this paper is structured as follows: Section \ref{sec_Disp}, presents the disparity map estimation. In Section \ref{sec_ipm}, IPM is described. In Section \ref{sec_SLT}, the SLT is described, and in Section \ref{sec_ESLT} the extended version of the SLT which is used in this paper is explained. In Section \ref{sec_Hough}, the Hough transform and its use in this paper are presented. In \ref{sec_Robust} robust fitting and its use in the algorithm is described. Section
\ref{sec_Masking} presents the details of the applied mask. Section 
\ref{sec_experimental_results}, illustrates the detection results, and in Section \ref{sec_conclusion}, the paper is concluded.
\section{Methodology}
\label{sec_methodology}

\subsection{Disparity Map Estimation}
\label{sec_Disp}
Computer stereo vision is commonly utilized to supply self-driving cars with 3D information \cite{fan2019real}. The state-of-the-art stereo vision methods can be categorized as either traditional \cite{Ihler2005, Tappen2003, Boykov2001, Fan2018a,  Hirschmuller2008, Mozerov2015} or convolutional neural network (CNN)-based \cite{Luo2016, Zagoruyko2015, Zbontar2015, Chang2018, Zhou2017}.

The traditional stereo vision algorithms are generally classified as local, global and semi-global \cite{Fan2018}. The local algorithms typically compare a collection of target image blocks with a chosen reference image block, and the shifting distances corresponding to the lowest cost are then determined as the best disparities \cite{Tippetts2016a}. Different from the local algorithms, the global algorithms generally formulate stereo matching as an energy minimization problem, which is subsequently addressed using Markov random field (MRF)-based optimizers \cite{sun2019active}, e.g., belief propagation (BP) \cite{Ihler2005} and graph cuts (GC) \cite{Boykov2001}. Semi-global matching (SGM) \cite{Hirschmuller2008} approximates the MRF inference by performing cost aggregation along all directions in the image, which greatly improves both the precision and efficiency of stereo matching. 

\begin{figure*}[!t]
	\begin{center}
		\centering
		\includegraphics[width=0.999\textwidth]{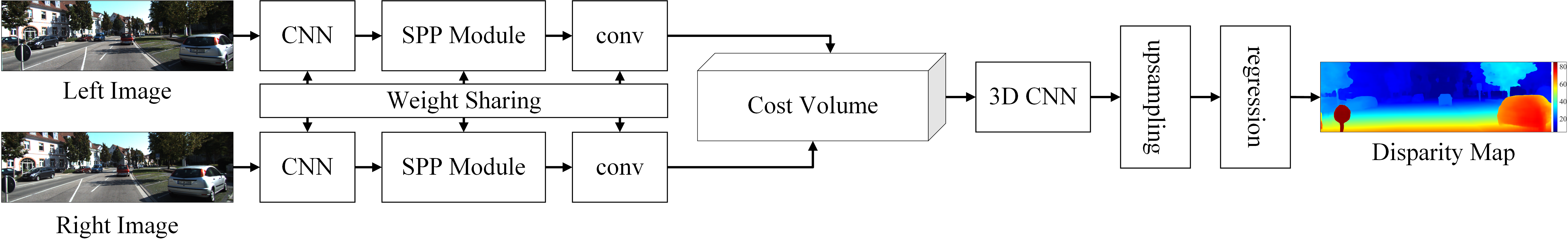}
		\caption{PSMNet architecture.}
		\label{fig.psmnet}
		\vspace{-2em}
	\end{center}
\end{figure*}

In recent years, CNN-based methods have achieved some very impressive disparity estimation and semantic segmentation results \cite{sun2019rtfnet}. These algorithms generally formulate disparity estimation as a binary classification problem and learn the probability distribution over all disparity values \cite{Luo2016}. For example, PSMNet \cite{Chang2018} generates the cost volumes by learning region-level features with different scales of receptive fields, and is regarded as one of the most accurate stereo vision methods for autonomous driving applications. Therefore, we utilize PSMNet in this paper to estimate disparity maps from stereo image pairs. The architecture of PSMNet is shown in Fig. \ref{fig.psmnet}, where SPP refers to spatial pyramid pooling. 

\subsection{Inverse Perspective Mapping}
\label{sec_ipm}
IPM is a commonly used technique to transform coordinate systems with different perspectives \cite{Jeong2016}. In this paper, we utilize IPM to map each image pixel $\textbf{p}=[u,v]^\top$ into a 3D world pixel $\textbf{P}=[X,Y,Z]^\top$. This can be straightforwardly realized using \cite{Bertozzi1998}: 
\begin{equation}
x(u,v)=h\cdot \text{ctg}\Big[(\theta-\alpha)+u\frac{2\alpha}{m-1}\Big]\cos\Big[(\gamma-\alpha)+v\frac{2\alpha}{n-1}\Big]
+l,
\label{eq.bev1}
\end{equation}
\begin{equation}
Z(u,v)=h\cdot \text{ctg}\Big[(\theta-\alpha)+u\frac{2\alpha}{m-1}\Big]\sin\Big[(\gamma-\alpha)+v\frac{2\alpha}{n-1}\Big]
+s,
\label{eq.bev2}
\end{equation}
where $h$ is the mounting height of the stereo rig, $\alpha$ is half of the camera angular aperture, $n\times m$, $\gamma$ is the yaw angle, which is assumed to be $0$ in this paper, and $\theta$ is the pitch angle.  

\begin{figure}[!t]
	\begin{center}
		\centering
		\includegraphics[width=0.49\textwidth]{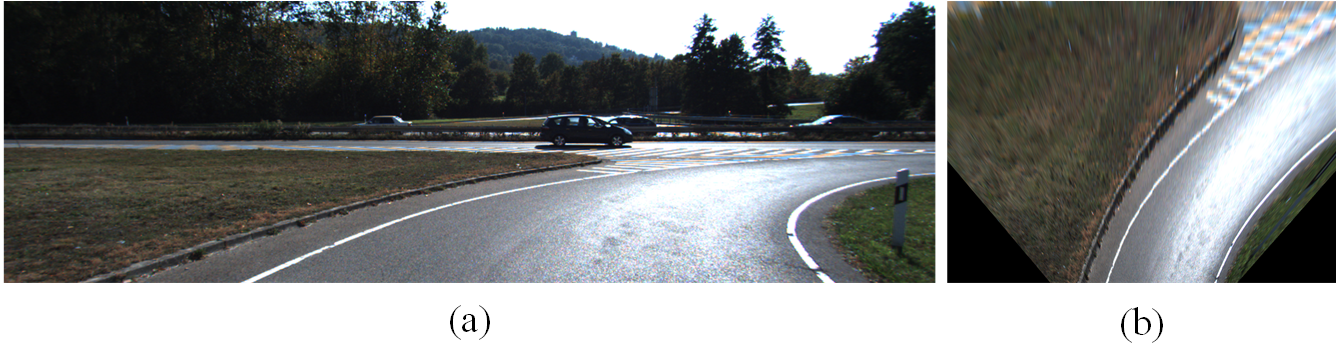}
		\caption{Bird's eye view transformation;: (a) original image; (b) bird's eye view image.}
		\label{fig.ipm}
	\end{center}
\end{figure}

In order to obtain the BEV map,  we first use v-disparity image analysis to segment the road region. The stereo rig roll angle is then estimated from the segmented road region using \cite{fan2018novel, Fan2019roaddamage, fan2019pothole}. Before estimating the pitch angle from the v-disparity image, we rotate the disparity map around the roll angle, which greatly improves the accuracy of pitch angle estimation. The pitch angle can be estimated using the following equation:
\begin{equation}
\theta=\arctan\Big(\frac{1}{f}\Big(  \frac{a_0}{a_1}+\frac{n-1}{2}        \Big)  \Big),
\label{eq.pitch_angle}
\end{equation}
where $f$ is the stereo camera focal length, and $a_0$ and $a_1$ are the coefficients of the road model. Plugging Eq. \ref{eq.pitch_angle} into Eqs. \ref{eq.bev1} and \ref{eq.bev2}, we can get the BEV map, as shown in Fig. \ref{fig.ipm}.

\subsection{SLT}
\label{sec_SLT}

The SLT is one of the most robust lane feature extraction algorithms. The SLT assumes that the lane markings have a lighter intensity compared to their backgrounds (e.g.,. asphalt). Thus, they have a DLD transition property. Consequently, for each pixel in a gray-scale image, the algorithm, compares the intensity value to the average of the pixels on the left-hand side within the range and the average of the pixels on the right-hand side within the range (on the same row) and, if the intensity value minus a threshold (to avoid incorrect estimations on the homogeneous regions) is higher than both of these values, that pixel is segmented as a lane feature point. The most important drawback of the SLT is that it does not supply any orientation information for the segmented feature points. However, many optimization algorithms use orientation for improved robustness and for efficiency in optimization.

\subsection{Extension to the SLT}
\label{sec_ESLT}

In our previous paper \cite{ozgunalp2014robust}, an extension to the SLT was proposed.  Kernels of the Sobel edge detector are applied to the input gray-scale image and the gradients of the pixels are calculated. Consequently, for each pixel, its intensity value and left average and right average are compared. Then, if the intensity of the pixel minus a threshold is higher than both values, the orientation of the pixel with the highest gradient on the left (left border) and the pixel with the highest gradient on the right (right border) are compared. If their orientations are similar as well (less than a 15 degree difference), the tested pixel is set to be a lane feature point and the orientation of that point is set to be the average of the orientation calculated for the left border and the orientation calculated for the left border. The pseudocode of the algorithm can be found in algorithm 1. For further information see \cite{ozgunalp2014robust}.

\begin{algorithm}[t]
\caption{Proposed feature extraction algorithm }
\begin{algorithmic} 
\STATE $F \leftarrow \emptyset$
\WHILE{$I_{p} \in Image Size $}
\IF{$I_{p}>Average_{R} +T_h \ and \ I_{p}>Average_{L} +T_h  $}
 
\STATE $\textbf{Calculate} \; \theta{max}(Iarea_{L}) \; and \; \theta max(Iarea_{R})$  
 
\IF{  $ \mid \theta{max}(Iarea_{L}) - \theta max(Iarea_{R}) \mid \ < \ \theta_{TH} $} 
 \STATE    $F_{Ip}\leftarrow 1$ 
 \ENDIF
 
\ENDIF
\ENDWHILE
\end{algorithmic}
\end{algorithm}

\subsection{1D Hough Transform and ROI Extraction}
\label{sec_Hough}

In the literature, many algorithms use the VP to improve accuracy of lane detection by first detecting the VP, which is global information, and then, detecting lanes which intersect at that point \cite{ma2018multiple}. For instance in \cite{Schnebele2015}, the first VP is detected, and then, lanes are detected using the 1D Hough transform (one parameter of a line is already known since it is known that the lines need to pass through this point).

In this paper, the lane detection is obtained in the BEV domain. However, a similar approach to VP-based lane detection algorithms can be adopted. Since lanes are assumed to be parallel to each other and in the BEV domain they remain parallel to each other, by creating a histogram of the orientations of the lane feature points, global lane orientation can be calculated. Conventionally, the Hough transform uses a 2D accumulator for detecting linear lines (a linear line has two parameters). In the polar Hough transform \cite{mukhopadhyay2015survey}, a Hough accumulator is created with the axes of  $\rho$ and $\theta$. Since, $\theta$ is already known, a 1D Hough accumulator with only parameter $\rho$ is created. Subsequently, linear lies are detected using a 1D Hough transform. The intersection of a line (on a feature point with known x and y coordinates and the line) and the bottom row of the image can be estimated with Eq. \ref{Eq_Hough}. This intersection point is calculated for each feature point and an accumulator cell is incremented based on the calculated $\rho$ \cite{fan2016faster}:
\begin{equation} 
\rho=x- \frac{ H-y }{ \tan(\theta ) },
\label{Eq_Hough}
\end{equation}

Thus, a 1D Hough accumulator is created with only one axis. Finally, the maximums of the two peak points are detected (one for the left lane and one for the right lane).

\subsection{Robust Fitting}
\label{sec_Robust}

In the previous section, lanes are detected. Although, this approach is robust and efficient, it defines lanes with linear lines. However, it is better to define lanes with a parabola, which supports curvature. Thus, ROIs with a fixed thickness (40 pixels) are defined with the already estimated $\rho$ and $\theta$. Then the feature points on this ROI are fitted to a parabola using robust fitting. This approach applies least squares fitting iteratively. Since the squares of the distances are considered, the least squares fitting is sensitive to outliers. However, robust fitting applies least squares fitting in the first iteration, then, in the next iteration gives a weight to each feature point (the closer to the estimated model, the higher the weight will be) and applies weighted least squares fitting. Thus, it is less sensitive to outliers.

\subsection{Masking}
\label{sec_Masking}

Although most of the noise is already dealt with, road markings, such as arrows on the road, can cause a problem, as these artifacts resemble lane markings. For instance an arrow has DLD property and its borders are also parallel to each other. Furthermore, they can be large in size, and while passing by, they can become very close to the camera. Thus, the algorithm must be able to distinguish these markings from, for example, dashed lanes. One distinction of these markings is that they do not appear consistently on consecutive frames, while lane markings can be detected consistently with a slight change in position. Thus, a mask is created based on the detection on the previous frame and applied to the next frame, largely preventing incorrect detections due to road markings.

\section{Experimental Results and Future Work}
\label{sec_experimental_results}

The proposed algorithm is tested using a video sequence from the KITTI dataset, and the detection ratio is found to be $91\%$. Example detection results, including failure cases, can be seen in Figure 3. The algorithm can detect lane markings robustly and accurately. The algorithm can easily deal with curved road markings, noise sources above the road (due to the applied mask calculated using the disparity map), and noise sources on the road. However, there are some cases that the algorithm fails to detect lanes correctly. These cases include when there is high curvature on the lanes and lots of noise on the ground in a single frame (see Figure 3(e)). Also, the algorithm fails when the input image saturates (see Figure 3(f)), and thus, lane markings are not visible in the frame. The only solution for these cases (especially in the case of saturation, since the lanes are not visible in the image) is to apply a tracking algorithm. Thus, we plan to apply a particle filter in future. 

\begin{figure}[t]
	\centering
\subfloat[]{\label{fig:}\includegraphics[width=0.34\linewidth , height=0.34\linewidth]{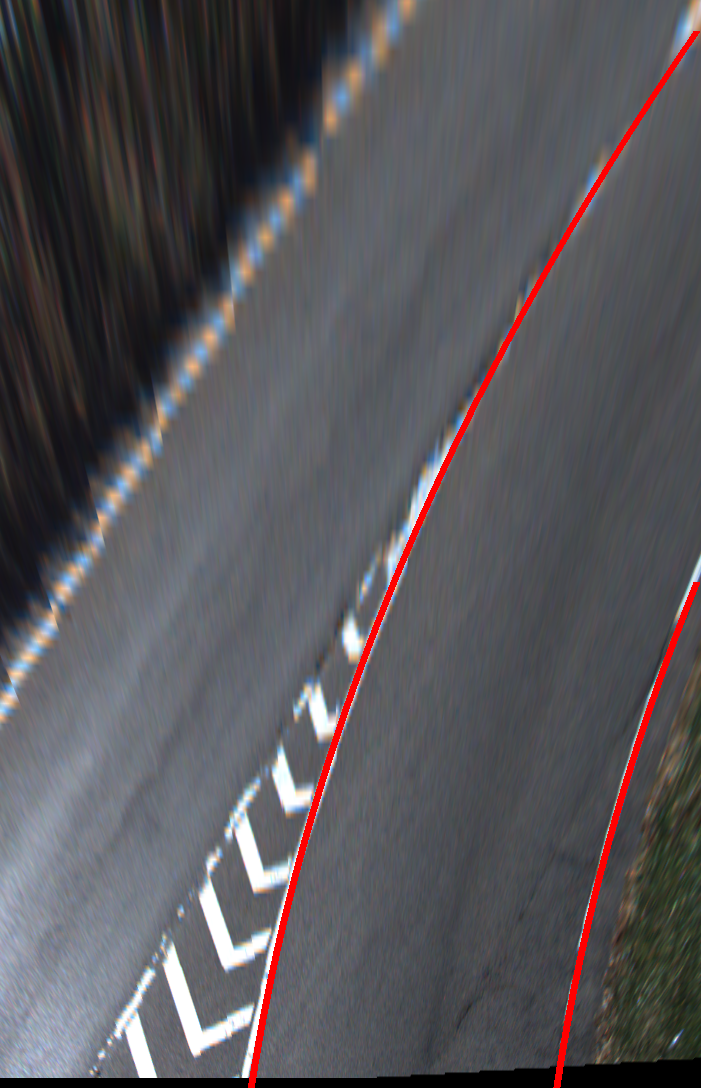}}  
 \qquad
\subfloat[]{\label{fig:}\includegraphics[width=0.34\linewidth , height=0.34\linewidth]{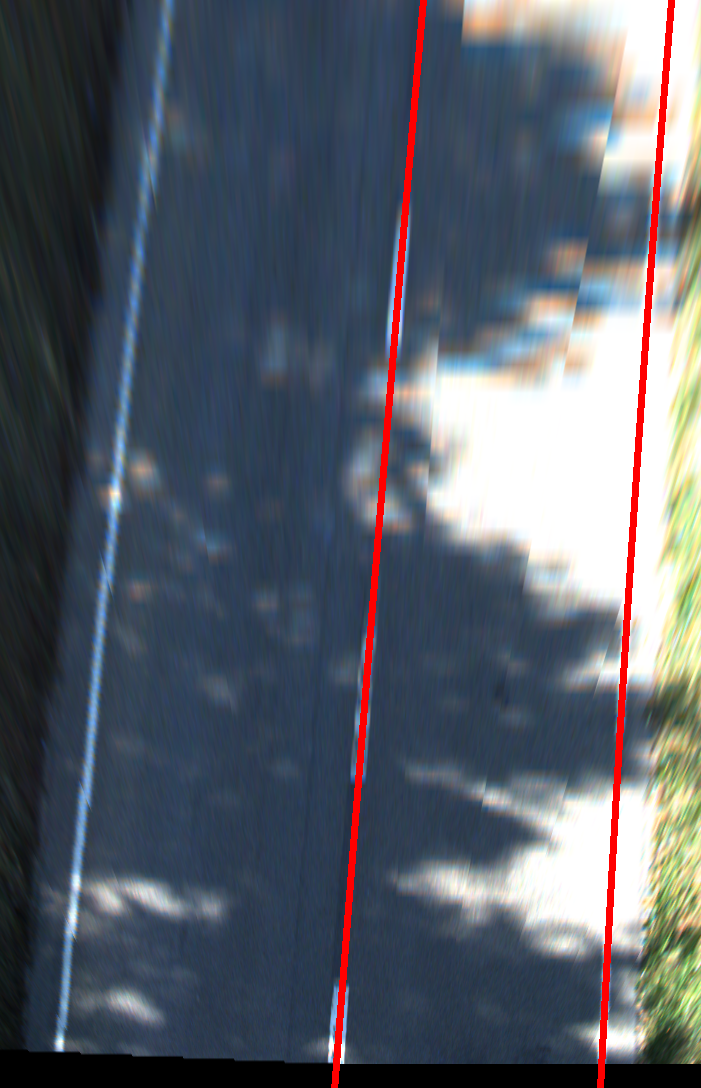}} 
\\
		
\subfloat[]{\label{fig:}\includegraphics[width=0.34\linewidth , height=0.34\linewidth]{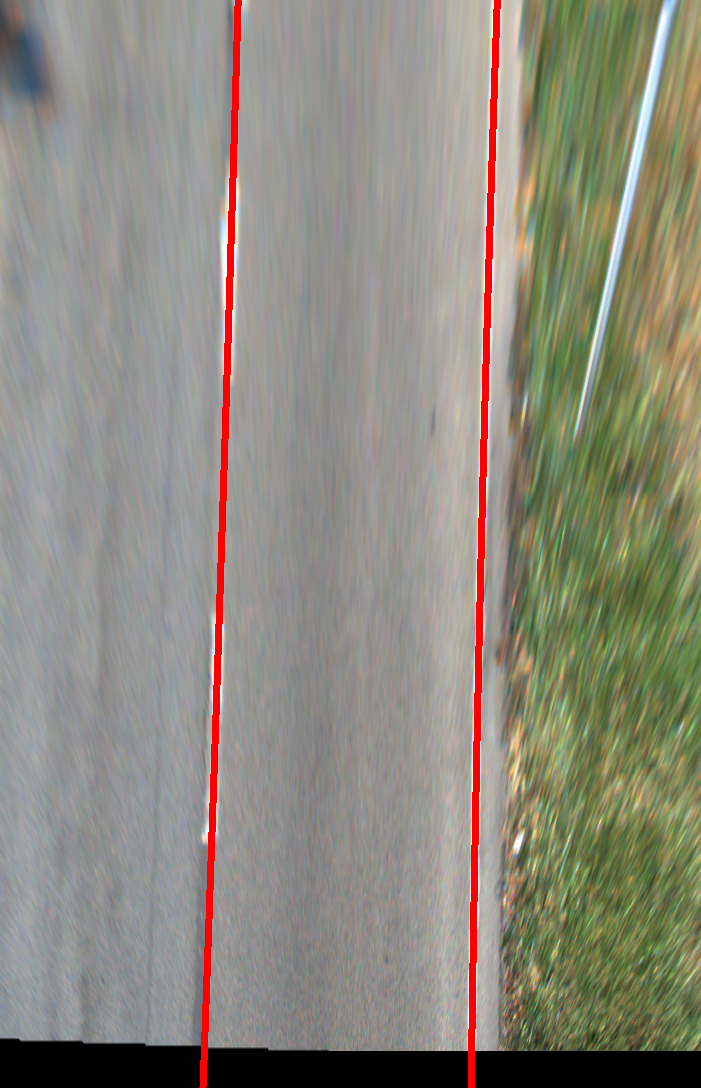}}
 \qquad
\subfloat[]{\label{fig:}\includegraphics[width=0.34\linewidth , height=0.34\linewidth]{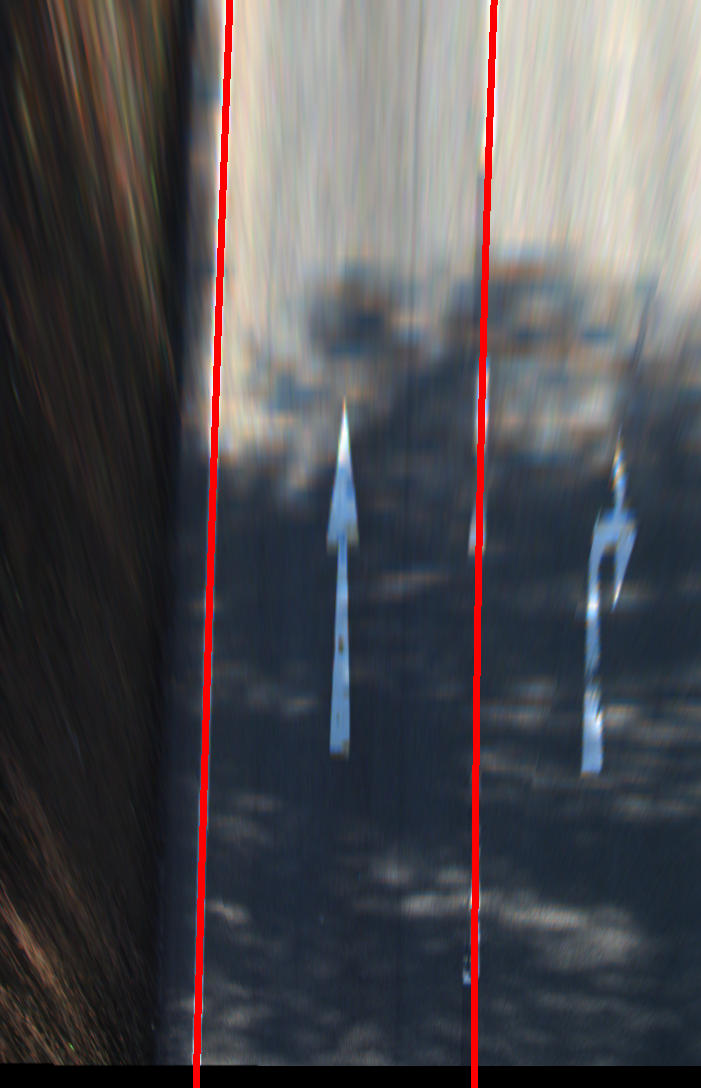}}  \\		
	
\subfloat[]{\label{fig:}\includegraphics[width=0.34\linewidth , height=0.34\linewidth]{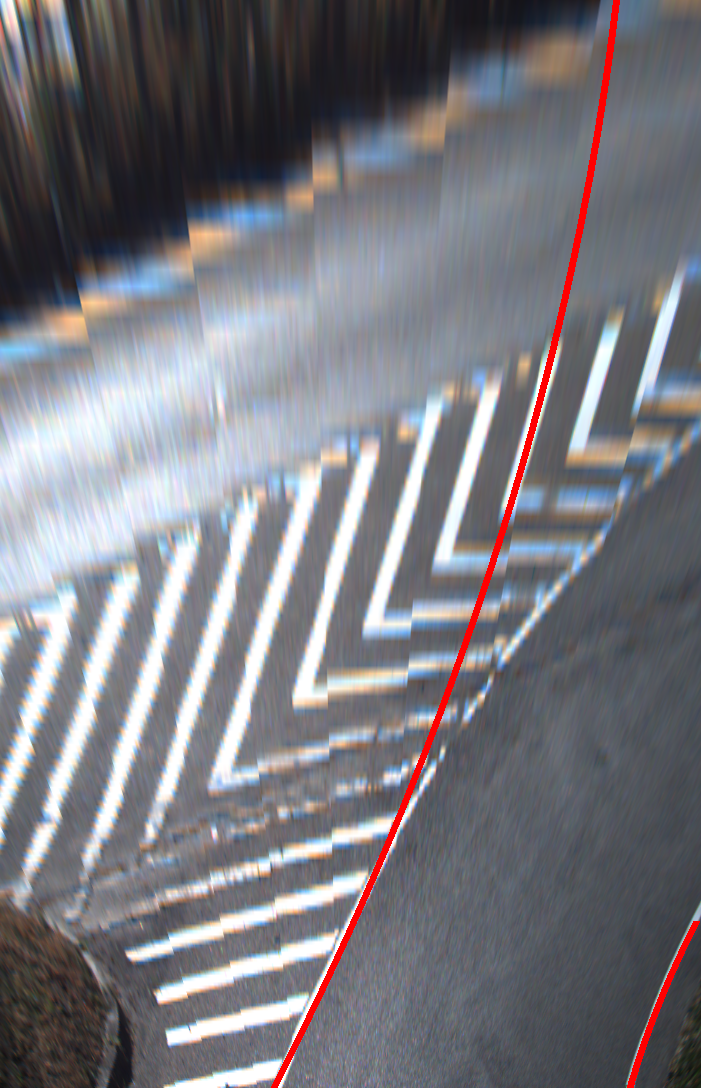}}
 \qquad
\subfloat[]{\label{fig:}\includegraphics[width=0.34\linewidth , height=0.34\linewidth]{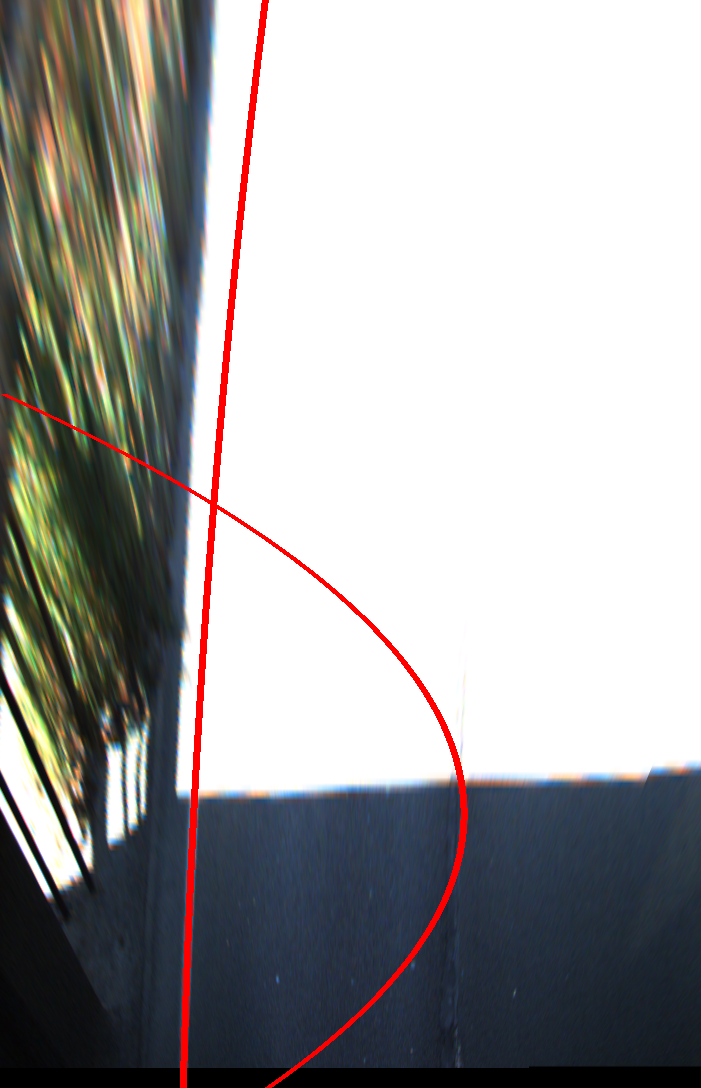}}

	\caption{Example lane detection results: (a), (b), (c), and (d) show the correct detections and (e) and (f) show incorrect detections.}

	 \label{fig8}
	\end{figure}

\section{Conclusion}
\label{sec_conclusion}

In this paper, a hybrid approach for lane detection was presented, where the road area was segmented using the dynamic programming on v-disparity map and the lanes are detected on the segmented road area. This robust lane detection approach is not sensitive to noise sources above the road, such as poles, trees, cars, sky, etc. The proposed lane detection algorithm can detect lanes robustly and accurately with a detection ratio of $91\%$.

\bibliographystyle{IEEEtran}

\begin{thebibliography}{10}
\providecommand{\url}[1]{#1}
\csname url@samestyle\endcsname
\providecommand{\newblock}{\relax}
\providecommand{\bibinfo}[2]{#2}
\providecommand{\BIBentrySTDinterwordspacing}{\spaceskip=0pt\relax}
\providecommand{\BIBentryALTinterwordstretchfactor}{4}
\providecommand{\BIBentryALTinterwordspacing}{\spaceskip=\fontdimen2\font plus
\BIBentryALTinterwordstretchfactor\fontdimen3\font minus
  \fontdimen4\font\relax}
\providecommand{\BIBforeignlanguage}[2]{{%
\expandafter\ifx\csname l@#1\endcsname\relax
\typeout{** WARNING: IEEEtran.bst: No hyphenation pattern has been}%
\typeout{** loaded for the language `#1'. Using the pattern for}%
\typeout{** the default language instead.}%
\else
\language=\csname l@#1\endcsname
\fi
#2}}
\providecommand{\BIBdecl}{\relax}
\BIBdecl

\bibitem{Fan2019}
R.~Fan, J.~Jiao, H.~Ye, Y.~Yu, I.~Pitas, and M.~Liu, ``Key ingredients of
  self-driving cars,'' \emph{arXiv:1906.02939}.

\bibitem{Bertozzi1998}
M.~Bertozzi and A.~Broggi, ``Gold: A parallel real-time stereo vision system
  for generic obstacle and lane detection,'' \emph{IEEE transactions on image
  processing}, vol.~7, no.~1, pp. 62--81, 1998.

\bibitem{fan2018real}
R.~Fan and N.~Dahnoun, ``Real-time stereo vision-based lane detection system,''
  \emph{Measurement Science and Technology}, vol.~29, no.~7, p. 074005, 2018.

\bibitem{Fan2018thesis}
R.~Fan, ``Real-time computer stereo vision for automotive applications,'' Ph.D.
  dissertation, University of Bristol, 2018.

\bibitem{neto2013real}
A.~M. Neto, A.~C. Victorino, I.~Fantoni, and J.~V. Ferreira, ``Real-time
  estimation of drivable image area based on monocular vision,'' in \emph{2013
  IEEE Intelligent Vehicles Symposium Workshops (IV Workshops)}.\hskip 1em plus
  0.5em minus 0.4em\relax IEEE, 2013, pp. 63--68.

\bibitem{schreiber2005single}
D.~Schreiber, B.~Alefs, and M.~Clabian, ``Single camera lane detection and
  tracking,'' in \emph{Proceedings. 2005 IEEE Intelligent Transportation
  Systems, 2005.}\hskip 1em plus 0.5em minus 0.4em\relax IEEE, 2005, pp.
  302--307.

\bibitem{ozgunalp2015lane}
U.~Ozgunalp and N.~Dahnoun, ``Lane detection based on improved feature map and
  efficient region of interest extraction,'' in \emph{2015 IEEE Global
  Conference on Signal and Information Processing (GlobalSIP)}.\hskip 1em plus
  0.5em minus 0.4em\relax IEEE, 2015, pp. 923--927.

\bibitem{ma2018multiple}
H.~Ma, Y.~Ma, J.~Jiao, M.~U.~M. Bhutta, M.~J. Bocus, L.~Wang, M.~Liu, and
  R.~Fan, ``Multiple lane detection algorithm based on optimised dense
  disparity map estimation,'' in \emph{2018 IEEE International Conference on
  Imaging Systems and Techniques (IST)}.\hskip 1em plus 0.5em minus 0.4em\relax
  IEEE, 2018, pp. 1--5.

\bibitem{ozgunalp2016multiple}
U.~Ozgunalp, R.~Fan, X.~Ai, and N.~Dahnoun, ``Multiple lane detection algorithm
  based on novel dense vanishing point estimation,'' \emph{IEEE Transactions on
  Intelligent Transportation Systems}, vol.~18, no.~3, pp. 621--632, 2016.

\bibitem{kluge1995deformable}
K.~Kluge and S.~Lakshmanan, ``A deformable-template approach to lane
  detection,'' in \emph{Proceedings of the Intelligent Vehicles' 95.
  Symposium}.\hskip 1em plus 0.5em minus 0.4em\relax IEEE, 1995, pp. 54--59.

\bibitem{broggi2010terramax}
A.~Broggi, A.~Cappalunga, C.~Caraffi, S.~Cattani, S.~Ghidoni, P.~Grisleri,
  P.~P. Porta, M.~Posterli, and P.~Zani, ``Terramax vision at the urban
  challenge 2007,'' \emph{IEEE Transactions on Intelligent Transportation
  Systems}, vol.~11, no.~1, pp. 194--205, 2010.

\bibitem{veit2008evaluation}
T.~Veit, J.-P. Tarel, P.~Nicolle, and P.~Charbonnier, ``Evaluation of road
  marking feature extraction,'' in \emph{2008 11th International IEEE
  Conference on Intelligent Transportation Systems}.\hskip 1em plus 0.5em minus
  0.4em\relax IEEE, 2008, pp. 174--181.

\bibitem{lopez2010robust}
A.~L{\'o}pez, J.~Serrat, C.~Canero, F.~Lumbreras, and T.~Graf, ``Robust lane
  markings detection and road geometry computation,'' \emph{International
  Journal of Automotive Technology}, vol.~11, no.~3, pp. 395--407, 2010.

\bibitem{otsu1979threshold}
N.~Otsu, ``A threshold selection method from gray-level histograms,''
  \emph{IEEE transactions on systems, man, and cybernetics}, vol.~9, no.~1, pp.
  62--66, 1979.

\bibitem{mccall2006video}
J.~C. McCall and M.~M. Trivedi, ``Video-based lane estimation and tracking for
  driver assistance: survey, system, and evaluation,'' 2006.

\bibitem{aubert1991autonomous}
D.~Aubert, K.~C. Kluge, and C.~E. Thorpe, ``Autonomous navigation of structured
  city roads,'' in \emph{Mobile Robots V}, vol. 1388.\hskip 1em plus 0.5em
  minus 0.4em\relax International Society for Optics and Photonics, 1991, pp.
  141--151.

\bibitem{ozgunalp2014robust}
U.~Ozgunalp and N.~Dahnoun, ``Robust lane detection \& tracking based on novel
  feature extraction and lane categorization,'' in \emph{2014 IEEE
  International Conference on Acoustics, Speech and Signal Processing
  (ICASSP)}.\hskip 1em plus 0.5em minus 0.4em\relax IEEE, 2014, pp. 8129--8133.

\bibitem{lim2009lane}
K.~H. Lim, K.~P. Seng, L.-M. Ang, and S.~W. Chin, ``Lane detection and
  kalman-based linear-parabolic lane tracking,'' in \emph{2009 International
  Conference on Intelligent Human-Machine Systems and Cybernetics},
  vol.~2.\hskip 1em plus 0.5em minus 0.4em\relax IEEE, 2009, pp. 351--354.

\bibitem{loose2009kalman}
H.~Loose, U.~Franke, and C.~Stiller, ``Kalman particle filter for lane
  recognition on rural roads,'' in \emph{2009 IEEE Intelligent Vehicles
  Symposium}.\hskip 1em plus 0.5em minus 0.4em\relax IEEE, 2009, pp. 60--65.

\bibitem{wang2004lane}
Y.~Wang, E.~K. Teoh, and D.~Shen, ``Lane detection and tracking using
  b-snake,'' \emph{Image and Vision computing}, vol.~22, no.~4, pp. 269--280,
  2004.

\bibitem{borkar2009robust}
A.~Borkar, M.~Hayes, and M.~T. Smith, ``Robust lane detection and tracking with
  ransac and kalman filter,'' in \emph{2009 16th IEEE International Conference
  on Image Processing (ICIP)}.\hskip 1em plus 0.5em minus 0.4em\relax IEEE,
  2009, pp. 3261--3264.

\bibitem{son2015real}
J.~Son, H.~Yoo, S.~Kim, and K.~Sohn, ``Real-time illumination invariant lane
  detection for lane departure warning system,'' \emph{Expert Systems with
  Applications}, vol.~42, no.~4, pp. 1816--1824, 2015.

\bibitem{fan2019real}
R.~Fan, J.~Jiao, J.~Pan, H.~Huang, S.~Shen, and M.~Liu, ``Real-time dense
  stereo embedded in a uav for road inspection,'' in \emph{Proceedings of the
  IEEE Conference on Computer Vision and Pattern Recognition Workshops}, 2019,
  pp. 0--0.

\bibitem{Ihler2005}
E.~T. Ihler, J.~W.~F. Iii, and A.~S. Willsky, ``Loopy belief propagation:
  Convergence and effects of message errors,'' 2005.

\bibitem{Tappen2003}
Tappen and Freeman, ``Comparison of graph cuts with belief propagation for
  stereo, using identical mrf parameters,'' in \emph{Proc. Ninth IEEE Int.
  Conf. Computer Vision}, Oct. 2003, pp. 900--906 vol.2.

\bibitem{Boykov2001}
Y.~Boykov, O.~Veksler, and R.~Zabih, ``Fast approximate energy minimization via
  graph cuts,'' \emph{IEEE Transactions on pattern analysis and machine
  intelligence}, vol.~23, no.~11, pp. 1222--1239, Nov. 2001.

\bibitem{Fan2018a}
R.~Fan, Y.~Liu, M.~J. Bocus, L.~Wang, and M.~Liu, ``Real-time subpixel fast
  bilateral stereo,'' in \emph{2018 IEEE International Conference on
  Information and Automation (ICIA)}.\hskip 1em plus 0.5em minus 0.4em\relax
  IEEE, 2018, pp. 1058--1065.

\bibitem{Hirschmuller2008}
H.~Hirschmuller, ``Stereo processing by semiglobal matching and mutual
  information,'' \emph{IEEE Transactions on pattern analysis and machine
  intelli- gence}, vol.~30, no.~2, pp. 328--341, Feb. 2008.

\bibitem{Mozerov2015}
M.~G. Mozerov and J.~van~de Weijer, ``Accurate stereo matching by two-step
  energy minimization,'' vol.~24, pp. 1153--1163, 2015.

\bibitem{Luo2016}
W.~Luo, A.~G. Schwing, and R.~Urtasun, ``Efficient deep learning for stereo
  matching,'' in \emph{Proceedings of the IEEE Conference on Computer Vision
  and Pattern Recognition}, 2016, pp. 5695--5703.

\bibitem{Zagoruyko2015}
S.~Zagoruyko and N.~Komodakis, ``Learning to compare image patches via
  convolutional neural networks,'' in \emph{Proceedings of the IEEE conference
  on computer vision and pattern recognition}, 2015, pp. 4353--4361.

\bibitem{Zbontar2015}
J.~Zbontar and Y.~LeCun, ``Computing the stereo matching cost with a
  convolutional neural network,'' in \emph{Proceedings of the IEEE conference
  on computer vision and pattern recognition}, 2015, pp. 1592--1599.

\bibitem{Chang2018}
J.-R. Chang and Y.-S. Chen, ``Pyramid stereo matching network,'' in
  \emph{Proceedings of the IEEE Conference on Computer Vision and Pattern
  Recognition}, 2018, pp. 5410--5418.

\bibitem{Zhou2017}
C.~Zhou, H.~Zhang, X.~Shen, and J.~Jia, ``Unsupervised learning of stereo
  matching,'' in \emph{Proceedings of the IEEE International Conference on
  Computer Vision}, 2017, pp. 1567--1575.

\bibitem{Fan2018}
R.~Fan, X.~Ai, and N.~Dahnoun, ``Road surface {3D} reconstruction based on
  dense subpixel disparity map estimation,'' \emph{IEEE Transactions on Image
  Processing}, vol.~PP, no.~99, p.~1, 2018.

\bibitem{Tippetts2016a}
B.~Tippetts, D.~J. Lee, K.~Lillywhite, and J.~Archibald, ``Review of stereo
  vision algorithms and their suitability for resource-limited systems,''
  \emph{Journal of Real-Time Image Processing}, vol.~11, no.~1, pp. 5--25,
  2016.

\bibitem{sun2019active}
Y.~Sun, M.~Liu, and M.~Q.-H. Meng, ``{Active Perception for Foreground
  Segmentation: An RGB-D Data-Based Background Modeling Method},'' \emph{{IEEE
  Transactions on Automation Science and Engineering}}, vol.~16, no.~4, pp.
  1596--1609, Oct 2019.

\bibitem{sun2019rtfnet}
Y.~Sun, W.~Zuo, and M.~Liu, ``{RTFNet: RGB-Thermal Fusion Network for Semantic
  Segmentation of Urban Scenes},'' \emph{{IEEE Robotics and Automation
  Letters}}, vol.~4, no.~3, pp. 2576--2583, July 2019.

\bibitem{Jeong2016}
J.~Jeong and A.~Kim, ``Adaptive inverse perspective mapping for lane map
  generation with slam,'' in \emph{2016 13th International Conference on
  Ubiquitous Robots and Ambient Intelligence (URAI)}.\hskip 1em plus 0.5em
  minus 0.4em\relax IEEE, 2016, pp. 38--41.

\bibitem{fan2018novel}
R.~Fan, M.~J. Bocus, and N.~Dahnoun, ``A novel disparity transformation
  algorithm for road segmentation,'' \emph{Information Processing Letters},
  vol. 140, pp. 18--24, 2018.

\bibitem{Fan2019roaddamage}
R.~{Fan} and M.~{Liu}, ``Road damage detection based on unsupervised disparity
  map segmentation,'' \emph{IEEE Transactions on Intelligent Transportation
  Systems}, pp. 1--6, 2019.

\bibitem{fan2019pothole}
R.~{Fan}, U.~{Ozgunalp}, B.~{Hosking}, M.~{Liu}, and I.~{Pitas}, ``Pothole
  detection based on disparity transformation and road surface modeling,''
  \emph{IEEE Transactions on Image Processing}, vol.~29, pp. 897--908, 2020.

\bibitem{Schnebele2015}
\BIBentryALTinterwordspacing
E.~Schnebele, B.~F. Tanyu, G.~Cervone, and N.~Waters, ``Review of remote
  sensing methodologies for pavement management and assessment,''
  \emph{European Transport Research Review}, vol.~7, no.~2, p.~1, Mar. 2015.
  [Online]. Available: \url{http://dx.doi.org/10.1007/s12544-015-0156-6}
\BIBentrySTDinterwordspacing

\bibitem{mukhopadhyay2015survey}
P.~Mukhopadhyay and B.~B. Chaudhuri, ``A survey of hough transform,''
  \emph{Pattern Recognition}, vol.~48, no.~3, pp. 993--1010, 2015.

\bibitem{fan2016faster}
R.~Fan, V.~Prokhorov, and N.~Dahnoun, ``Faster-than-real-time linear lane
  detection implementation using soc dsp tms320c6678,'' in \emph{2016 IEEE
  International Conference on Imaging Systems and Techniques (IST)}.\hskip 1em
  plus 0.5em minus 0.4em\relax IEEE, 2016, pp. 306--311.

\end{thebibliography}
\balance


\end{document}